# Learning for Dynamic Bidding in Cognitive Radio Resources

Fangwen Fu[1], and Mihaela van der Schaar

### ABSTRACT

In this paper, we model the various wireless users in a cognitive radio network as a collection of selfish, autonomous agents that strategically interact in order to acquire the dynamically available spectrum opportunities. Our main focus is on developing solutions for wireless users to successfully compete with each other for the limited and time-varying spectrum opportunities, given the experienced dynamics in the wireless network. We categorize these dynamics into two types: one is the disturbance due to the environment (e.g. wireless channel conditions, source traffic characteristics, etc.) and the other is the impact caused by competing users. To analyze the interactions among users given the environment disturbance, we propose a general stochastic framework for modeling how the competition among users for spectrum opportunities evolves over time. At each stage of the dynamic resource allocation, a central spectrum moderator auctions the available resources and the users strategically bid for the required resources. The joint bid actions affect the resource allocation and hence, the rewards and future strategies of all users. Based on the observed resource allocation and corresponding rewards from previous allocations, we propose a best response learning algorithm that can be deployed by wireless users to improve their bidding policy at each stage. The simulation results show that by deploying the proposed best response learning algorithm, the wireless users can significantly improve their own performance in terms of both the packet loss rate and the incurred cost for the used resources.

Manuscript submitted: August 28, 2007. The authors are with the University of California Los Angeles (UCLA), Dept. of Electrical Engineering (EE), 66-147 Engineering IV Building, 420 Westwood Plaza, Los Angeles, CA, 90095-1594, USA, Phone:+1-310-825-5843, Fax:+1-310-206-4685, Email: {fwfu, mihaela}@ee.ucla.edu.
[1] Corresponding author. Address and contact information: see above.

# I. INTRODUCTION

One vision for emerging cognitive radio networks assumes that certain portions of the spectrum will be opened up for secondary users[2] (SUs), which can autonomously and opportunistically share the spectrum once primary users (PUs) are not active [1][2][3]. Prior work has mainly focused on addressing two main challenges. The first problem is the detection of spectrum opportunities ("holes") that can be used by SUs for transmission [4][5][6]. The second challenge is developing resource allocation solutions for the efficient usage of the detected spectrum holes among autonomous wireless users [7][8]. In this paper, our main focus will be on addressing a third challenge. We focus on developing solutions that can be employed by the SUs to improve their performance in the cognitive radio network. Specifically, we aim at investigating how delay-sensitive applications (e.g. multimedia applications) can efficiently forecast their future utility impact, and then determine their resource requirements and associated transmission strategies over time, based on information about the available spectrum opportunities, their source and channel characteristics, and interactions with the other competing users. We concentrate in this paper on delay-sensitive data transmission applications such as real-time multimedia streaming, videoconferencing, virtual reality games etc., because they can most benefit from the additional bandwidth resources provided by emerging cognitive radio networks. Hence, if relaxed FCC regulations are to be extended to large portions of the spectrum, improved support for delay-sensitive, high-bandwidth applications should be provided.

Solving all these aforementioned challenges is important for the proliferation of both cognitive radio networks and applications that use these new networks. To address the first challenge, we rely in this paper on existing methods for identifying the spectrum holes. Concerning solutions for spectrum management, several centralized and distributed solutions already exist [11][19][36][37], including our prior research in [10][28]. While our main interest in this paper is not on developing new solutions for addressing this spectrum management challenge, we present a general framework for the *dynamic* interaction in the wireless resource management, since this will enable us to investigate the multi-user interaction among delay-sensitive, time-varying multimedia applications in the cognitive radio network in a universal setting. Our main concern in developing a successful spectrum management solution for emerging cognitive radio networks is not only to maximize the network utilization, but also to take into account the "self-interested" behavior of individual users/applications that may try to selfishly influence the resource management. In this way, we recognize the informational constraints and distributed nature of the multi-user resource allocation problem, where the private information of each user is not known by the system or other users.

---

[2] The secondary users/applications are envisioned in this paper to be a single transmitter-receiver pair.

In this paper, we assume that multi-user spectrum access in cognitive radio networks is similar to that currently deployed wireless networks, e.g. WLAN, etc [14]. We model the dynamic spectrum management as a stochastic, repeated interaction among autonomous SUs aiming at maximizing their own utilities. Unlike the previous works on resource management, our main focus is on discussing how users can adapt, predict and learn their resource requirements (also referred to as resource bids) and associated transmission strategies given the experienced "dynamics". In the cognitive radio network, these dynamics can be categorized into two types: one is the *disturbance due to the "environment"* and the other is the *impact caused by competing users*. The disturbance due to the environment results from variations (uncertainties) of the wireless channels or source (e.g. multimedia) characteristics. For example, the stochastic behavior of the primary users, the time-varying channel conditions experienced by the SUs and the time-varying source traffic that needs to be transmitted by the SUs can be considered as environment disturbances. These types of dynamics are generally modeled as stationary processes. For instance, the usage of each channel by the primary users can be modeled as a two-state Markov chain with ON (the channel is used by PUs) and OFF (the channel is available for the SUs) states [12]. The channel conditions can be modeled using a finite state Markov model [33]. The packet arrival of the source traffic can be modeled as a Poisson process[3] [16].

Conventionally, wireless stations have only considered these environment disturbances when adapting their cross-layer strategies [17] for delay-sensitive transmission. The other type of dynamics - the impact from competing users, which is due to the non-collaborative, autonomous and strategic SUs in the network transmitting their traffic - is less well studied in wireless communication networks.

The goal of this paper is to provide solutions and associated metrics that can be used by an autonomous SU to analyze and predict the outcome of various dynamic interactions among competing SUs in dynamic multi-user communication systems and, based on this forecast, adapt and optimize its transmission strategy. In our considered cognitive radio network, the SUs are modeled as rational and strategic (i.e. they may manipulate their declaration of the required resources. We model the spectrum management as a stochastic repeated interaction/game [29] in which the SUs simultaneously and repeatedly make their own resource bids. The competition for the dynamic resources is assisted by a central coordinator (similar to that in existing wireless LAN standards such as 802.11e HCF [18]). We refer to this coordinator as the central spectrum moderator (CSM). The role of the CSM is to allocate resources to the SUs based on pre-determined utility maximization rule[4].

---

[3] Other packet arrival models can also been used in our proposed framework.

In this paper, to explicitly consider the strategic behavior of the autonomous SUs and the informationally-decentralized nature of the competition for wireless resources, we assume that the CSM deploys an auction mechanism for dynamically allocating resources. Auction theory has been extensively studied in economics [25] and it has also been recently applied to network resource allocation [9][10][11][26]. Note that the role of the CSM [5] in our resource management game for cognitive radio networks will be kept minimal. Unlike alternative existing solutions [28], the CSM will not require knowledge of the private information of the users and will not perform complex computations for deciding the resource allocation. Its only role will be the implementation of the spectrum etiquette rules as in [13], and ensuring that the available spectrum holes are auctioned among users. In order to capture the network dynamics, we allow the CSM to *repeatedly* auction the available spectrum opportunities based on the PUs' behaviors. Meanwhile, each SUs is allowed to strategically adapt its bidding strategy based on information about the available spectrum opportunities, its source and channel characteristics, and impacts of the other SUs bidding actions.

Using this general stochastic wireless allocation framework, the key focus of this paper is to develop a learning methodology for SUs to improve their policies for playing the auction game, i.e. the policies for generating the bids for the available resources. Specifically, during the repeated multi-user interaction, the SUs can observe partial historic information of the outcome of the auction game, through which the SUs can estimate the impact on their future rewards and then adopt their best response in order to effectively compete for the channel opportunities. The estimation of the impact on the expected future reward can be performed using different types of interactive learning [24]. In this paper, we focus on reinforcement learning [23][38] because this allows the SUs to improve their bidding strategy based only on the knowledge of their own past received payoffs, without knowing the bids or payoffs of the other SUs. Our proposed best response learning algorithm is inspired from the Q-learning for the single agent interacting with environment. Unlike the Q-learning, the proposed best response learning explicitly considers the interactions and coupling among SUs in this multi-user cognitive radio network. By deploying the best response learning algorithm, the SUs can strategically predict the impact of current actions on future performance and then optimally make their bids.

---

[4] Other fairness rules can also be deployed in the CSM such as air-time fairness, utility-based fairness, etc. [17]

[5] It should be noted that this approach can also allow for multiple CSMs to manage the spectrum, by dividing their responsibilities fairly, e.g. based on their geo-location [34] or frequency band in which they are operating, or by competing against each other for the number of SUs that will associated with them.

The paper is organized as follows. In Section II, we describe a system model for cognitive radio network and a general model for the resource competition among the SUs. In Section III, we propose a stochastic framework to model the multi-user interactions in the cognitive radio network. In Section IV, we propose a best response learning approach for the SUs to predict their future rewards based on the observed historic information. In Section V, we present the simulation results, followed by the conclusions in Section VI.

## II. DYNAMIC MULTI-USER WIRELESS INTERACTION FRAMEWORK

We consider a spectrum consisting of $N$ channels, each indexed by $j \in \{1,...,N\}$. The $N$ wireless channels are originally licensed to a primary network (PN) whose users (i.e. PUs) exclusively access the channels. The communications of the PUs are assumed to follow the synchronous slot structure. The time slot has length $\Delta T$ in seconds. At each time slot, each channel is assumed to be in one of the following two states: ON (this channel is currently used by the PUs) or OFF (this channel is not used by the PUs and hence is an opportunity for the SUs to use). Within each time slot, the channel is only OFF or ON [14]. At time slot $t \in \mathbb{N}$, the availability of each channel $j$ is denoted by $y_j^t \in \{0,1\}$ with $y_j^t$ being 0 if the channel is in ON state, and being 1 if it is in OFF state. The channel availability profile for the $N$ channels is represented by $\boldsymbol{y}^t = \left[ y_1^t,...,y_N^t \right]$. This can be characterized using a Spectrum Opportunity Map [14]. Note that our proposed framework is not restricted to a cognitive radio network which has licensed users, i.e. PUs. Our solution can also be applied to cognitive radio networks using multiple ISM bands.

We consider the situation in which $M$ $(M \geq N)$ autonomous SUs transmitting delay-sensitive bitstreams compete for the spectrum opportunities in these $N$ channels. Each SU is indexed by $i \in \{1,...,M\}$. Our goal is to develop a general mechanism for coordinating the competition for spectrum opportunities among SUs and, more importantly, to understand the competitive interactions among SUs across time, thereby enabling SUs to improve their strategies for playing the repeated resource management game based on their past interactions with other SUs.

As in [18], we assume that a polling-based medium access protocol is deployed in the secondary network, which is arbitrated by a CSM. The polling policy is changed only at the beginning of every time slot. For simplicity, we assume that each SU can access a single channel and that each channel can be accessed by a single SU within the time slot. The SUs can switch the channels only when crossing time slots. Note that this simple medium access model used for illustration in this paper can be easily extended to more sophisticated cognitive radio models [15], where each SU can simultaneously access multiple channels or the channels are being shared by multiple SUs etc.

We assume that the CSM is aware of the channel availability profile $\boldsymbol{y}^t$ and allocates (through polling the SUs) those channels with $y_j^t = 1$ to the SUs. To efficiently allocate the available resources (opportunities), the CSM needs to collect information about the SUs [28]. However, as mentioned in Section I, in a wireless network, the information is decentralized, and thus, the information exchange between the SUs and CSM needs to be kept limited due to the incurred communication cost. On the other hand, the SUs competing with each other are selfish and strategic, and hence, the information they hold is private and may not be shared with each other. Therefore, one of our key interests in this paper is to determine what information should be exchanged between SUs and CSM and how this information should be exchanged. In the subsequent sections, we present an auction mechanism for dynamically coordinating the interactions among SUs and subsequently discuss the bidding strategy for the SUs.

## A. Resource Auction Mechanism

We model the multi-user wireless resource allocation as an auction for spectrum opportunities held by the CSM during each time slot. First, the CSM announces the auction by broadcasting the channel opportunity profile $\boldsymbol{y}^t$. The SUs receive the announcement and calculate the bid vector $\boldsymbol{b}_i^t = \left[ b_{i1}^t, \ldots, b_{iN}^t \right] \in \mathbb{R}^N$ based on the announced information and their own private information about the environment they experience, which is discussed in details in Section III. Subsequently, each SU submits the bid vector to the CSM. After receiving the bid vectors from the SUs, the CSM computes the channel allocation $\boldsymbol{z}_i^t = \left[ z_{i1}^t, \ldots, z_{iN}^t \right] \in \{0,1\}^N$ for each SU $i$ based on the submitted bids. To compel the SUs to declare their bids truthfully [32], the CSM also computes the payment $\tau_i^t \in \mathbb{R}_-$ that the SUs have to pay for the use of resources during the current stage of the game. The negative value of the payment means the absolute value that SU $i$ has to pay the CSM for the used resources. The auction result is then transmitted back to the SUs which can deploy their transmission strategies in different layers and send data over the assigned channel. After the data transmission, another auction starts at the next time slot $t+1$. The information exchange between the CSM and SUs is illustrated in Figure 1. The computation of the channel allocation $z_i^t$ and payment $\tau_i^t$ is described as follows.

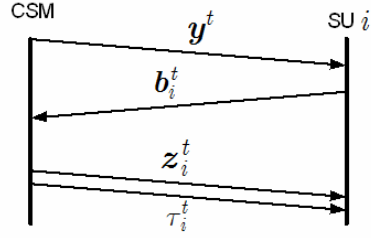

Figure 1.   Information exchange between the CSM and SU $i$

After each SU submits the bid vector, the CSM performs two computations: (i) channel allocation and (ii) payment computation. Note that most existing multi-user wireless resource allocation solutions can be modeled as such repeated auctions for resources. If the resources are priced or the users may lie about their resource needs, taxes associated with the resource usage will need to be imposed [20]. Otherwise, these taxes can be considered to be zero throughout the paper.

We denote the channel allocation matrix $Z^t = [z_{ij}^t]_{M \times N}$ with $z_{ij}^t$ being 1 if channel $j$ is assigned to SU $i$ , otherwise 0. The feasible set of channel assignments is denoted as $\mathcal{Z}^t = \{Z^t \mid \sum_{i=1}^{M} z_{ij}^t = y_j^t, \forall j, \sum_{j=1}^{N} z_{ij}^t \leq 1, \forall i, z_{ij}^t \in \{0,1\}\}$ . The channel allocation matrix without the presence of SU $i$ is denoted $Z_{-i}^t = [z_{kj}^t]_{(M-1) \times N}$ and the corresponding feasible set is $\mathcal{Z}_{-i}^t = \{Z_{-i}^t \mid \sum_{k=1,k \neq i}^{M} z_{kj}^t = y_j^t, \forall j, \sum_{j=1}^{N} z_{kj}^t \leq 1, \forall k \neq i, z_{kj}^t \in \{0,1\}\}$ , where $-i = \{1,...,i-1,i+1,...,M\}$ . During the first phase, the CSM allocates the channels to SUs based on its adopted fairness rule, e.g. maximizing the total "social welfare"[6]:

$$Z^{t,opt} = \arg \max_{Z^t \in \mathcal{Z}^t} \sum_{i=1}^{M} \sum_{j=1}^{N} z_{ij}^t b_{ij}^t . \tag{1}$$

If the resources are priced, we will consider in this paper, for illustration, a second price auction mechanism [25][32] for determining the tax that needs to be paid by SU $i$ based on the above optimal channel assignment $Z^{t,opt} = [z_{ij}^{t,opt}]_{M \times N}$ . This tax equals:

$$\tau_i^t = \sum_{k=1,k \neq i}^{M} \sum_{j=1}^{N} z_{kj}^{t,opt} b_{kj}^t - \max_{Z_{-i}^t \in \mathcal{Z}_{-i}^t} \sum_{k=1,k \neq i}^{M} \sum_{j=1}^{N} z_{kj}^t b_{kj}^t . \tag{2}$$

Although the optimization problems in Eqs. (1) and (2) are discrete optimizations, they can be efficiently solved using linear programming or heuristic algorithms [27]. Note that when $N = 1$, the generalized auction mechanism presented above becomes the well-known second price auction [25].

---

[6] Note that other fairness solutions than maximizing the social welfare could be adopted and this will not influence our proposed solution.

## B. Bidding Policy for SUs

As discussed in the introduction section, the SUs experience a dynamic environment due to the time-varying source characteristics and channel conditions as well as the channel opportunities. The variations in the environment result in different transmission strategies (e.g. cross-layer optimization [17]), as well as different bid vectors for the available channels. The environment experienced by an SU can be characterized by its current "state" which will be discussed in Section III. The bidding policy of each SU generates the bid vector based on its current "state". After receiving the allocated channels, the SUs adapt their strategy for transmitting the packets and transition to new states in order to play the auction game in the next time slot. The conceptual overview of the multi-SUs interactions in the repeated auctions is illustrated in Figure 2.

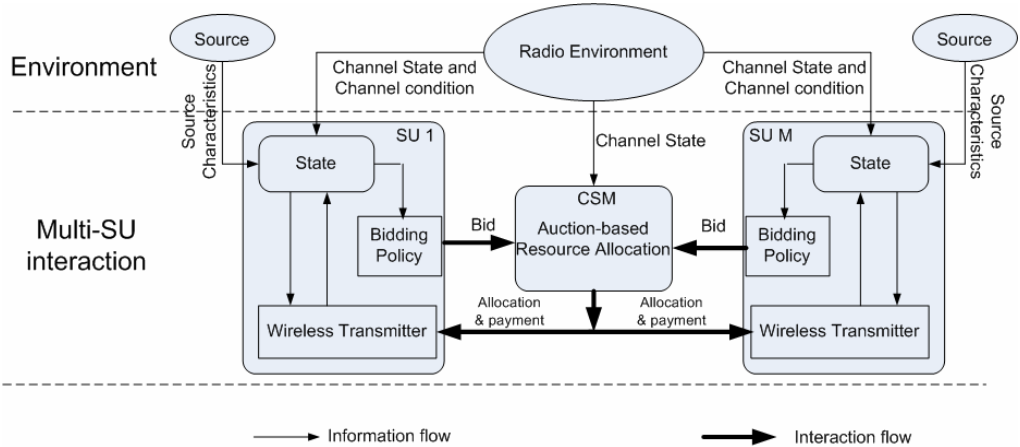

Figure 2.   Conceptual overview of the multi-SU interaction in cognitive radio network

The key questions to determine the bidding policy in the repeated auctions among the SUs are: (i) what kind of environment (i.e. current state) each SU experiences in each time slot in the dynamic cognitive radio network; (ii) during the repeated competition, how the interactions among the SUs are modeled; and (iii) how the SUs forecast the impact of the bids made in the current time on the future performance.

To overcome the above addressed problems, we present in the next section a stochastic framework for modeling the dynamic interaction among users.

## III. STOCHASTIC MODEL FOR SUs INTERACTION

The SUs bid for the available channels in the repeated auctions given their dynamically changing environment which they experience. The evolution of SUs' interactions across the various time slots can be modeled as a stochastic interaction. From [21], we know that the Markov Decision Process (MDP) is a decision problem for one agent (corresponding to the SU in this paper) in an (unknown) environment. When multiple agents compete or interact with each other in such an environment, this becomes a stochastic game [22][29] problem (i.e. n-agent MDP). In the stochastic game, the time is discrete and at each stage, every agent has its own state and its own action space for that state. The time slot corresponds to the "stage" commonly used in the stochastic game. In the remainder of the paper, we use the time slot and stage interchangeably. The agents choose their own actions independently and simultaneously at each time slot. Next, they receive their rewards and transit to the next states. It is worth noting that the reward received by each agent, and state transition also depend on other agents' states and actions.

Formally, a stochastic game is a tuple $(\mathcal{I}, \mathcal{S}, \mathcal{B}, P, \mathcal{R})$, where $\mathcal{I}$ is the set of agents (SUs), i.e. $\mathcal{I} = \{1, \dots, M\}$, $\mathcal{S}$ is the set of state profiles of all SUs, i.e. $\mathcal{S} = \mathcal{S}_1 \times \cdots \times \mathcal{S}_M$ with $\mathcal{S}_i$ being the state set of SU $i$, and $\mathcal{B}$ is the joint action space $\mathcal{B} = \mathcal{B}_1 \times \cdots \times \mathcal{B}_M$, with $\mathcal{B}_i$ [7] being the action (bidding) set available for SU $i$ to play the game. $P$ is a transition probability function defined as a mapping from the current state profile $s \in \mathcal{S}$, corresponding joint actions $b \in \mathcal{B}$ and the next state profile $s^{'} \in \mathcal{S}$ to a real number between 0 and 1, i.e. $P: \mathcal{S} \times \mathcal{B} \times \mathcal{S} \mapsto [0,1]$. $\mathcal{R}$ is a reward vector function defined as a mapping from the current state profile $s \in \mathcal{S}$ and corresponding joint actions $b \in \mathcal{B}$ to an $M$-dimensional real vector with each element being the reward to a particular agent, i.e. $\mathcal{R}: \mathcal{S} \times \mathcal{B} \mapsto \mathbb{R}^M$.

Recall $-i = \{1, \dots, i-1, i+1, \dots, M\}$. The state profile $s \in \mathcal{S}$ can be sometimes rewritten as $s = (s_i, s_{-i})$ to distinguish the state of agent $i$ and the states of other agents. Similarly, the joint action $b \in \mathcal{B}$ can also be represented as $b = (b_i, b_{-i})$. In the subsequent sections, we specify the elements of the stochastic game model for the interactions among the SUs in the considered cognitive radio network.

### A. Definition of SU States

As discussed in the introduction, each SU needs to cope with two types of "uncertainties" in terms of available resources: disturbances from the environment and interactions with other SUs. The environment is characterized by the packet arrivals from the source (i.e. source/traffic characterization) connected with the transmitter, the spectrum opportunities released by PUs and the channel condition. Next, we will illustrate how these disturbances can be modeled. However, note that other models of the environment existing in the literature can be adopted. The use of a specific model will only affect the performance of the proposed solution, and not the general framework for multi-user interaction proposed in this work.

---

[7] Note that the action set may depend on the state the SU is in. For simplicity, we assume the actions set are the same for all the states the SU lies in.

For illustration, we assume that each SU $i$ maintains a buffer with limited size $B_i$, which can be interpreted as a time window that specifies which packets are considered for transmission at each time based on their delay deadlines. Expired packets are dropped from the buffer. This model has been extensively used for delay-sensitive data transmission, e.g. leaky bucket model for video transmission [35]. The number of packets in the buffer at time slot $t$ is denoted as $v_i^t$ ( $0 \leq v_i^t \leq B_i$ ). We assume that the packets arrive from the source at the beginning of each time slot, i.e. $v_i^t$ is updated only at the beginning of a time slot.

The availability profile of the $N$ channels, $\boldsymbol{y}^t$, is assumed to be known by the CSM at the beginning of each stage and also announced by the CSM to all the participating SUs. The condition of channel $j$ experienced by SU $i$ is represented by the Signal-to-Noise Ratio (SNR) and it is denoted as $c_{ij}^t$ in dB. If the spectrum opportunity in channel $y_j^t$ is 0, we set $c_{ij}^t$ equal to $-\infty$ which means that the channel is unavailable to SUs at that time. We assume that the channel condition of each channel can be represented by a set of discrete SNR values $\{\sigma_{i1},...,\sigma_{iK}\}$ when $y_j^t = 1$ ; otherwise the channel condition is $-\infty$ . Note that the number of discrete SNR values, $K$ , can be determined by SU $i$ by trading-off the complexity (larger $K$ leads to a larger state space) and the resulting impact on the performance. We define a state for each channel $j$ as $w_{ij}^t = (y_j^t, c_{ij}^t)$ . Hence, $w_{ij}^t$ can only take values in $\{(0, -\infty),(1,\sigma_{i1}),...,(1,\sigma_{iK})\}$ . The total number of possible $w_{ij}^t$ is $(K+1)$ . The channel state profile is $\boldsymbol{w}_i^t = \left(w_{i1}^t,...,w_{iN}^t\right)$ . The models for the packet arrivals, the transition of channel availability and channel condition across time slots are described in Section III.B.

To model the dynamics experienced by SU $i$ at time $t$ in the cognitive radio network, we define a "state" $s_i^t = (v_i^t, \boldsymbol{w}_i^t) \in \mathcal{S}_i$ , which encapsulates the current buffer state as well as the state of each channel. $\mathcal{S}_i$ is the set of possible states [8] . The total number of possible states for SU $i$ equals $|\mathcal{S}_i| = (B_i + 1) \times (K+1)^N$ . We will show later in this paper that the state information is sufficient for SU $i$ to compete for resources (make bid vector) at the current time.

### B. Environment Modeling and State Transition

In this section, to simplify the description, we use simple models to capture the uncertainties in the

---

environment of the cognitive radio network. Note though that our proposed stochastic game framework can also be extended to more sophisticated models for the cognitive radio network. Based on the environment model, we further derive the state transition for the stochastic multi-user interaction.

*(1)   Model for Channel State Evolution*

The opportunity of each channel $j$, $y_j^t$, is independently modeled by a two state Markov chain [15]. The probabilities of channel $j$ are $p_j^{NF}$ for the transition from the ON state to OFF state and $p_j^{FN}$ for the transition from the OFF state to ON state. We assume that the transition probabilities, $p_j^{NF}$ and $p_j^{FN}$, $\forall j$ are known to all SUs. For example, the CSM can predict these probabilities and announce them to the participating SUs. Each channel experienced by SU $i$ is characterized by its SNR, denoted by $c_{ij}^t$. As discussed previously, the value of the SNR is only taken from a finite set $\{\sigma_{i1},...,\sigma_{iK}\}$ when $y_j^t = 1$. As shown in [30], when $y_j^t = 1$, the channel condition (in terms of SNR) can also be modeled as a finite-state Markov chain, where the transition from channel condition $\sigma_{il}$ at time $t$ to channel condition $\sigma_{ik}$ at time $t+1$ takes place with probability $p_{ij}^{l \rightarrow k}$. These transitions probabilities can be easily estimated by SU $i$, by repeatedly interacting with the channel. Hence, the state transition from $w_{ij}^t = (1, \sigma_{il})$ to $w_{ij}^{t+1} = (1, \sigma_{il})$ has probability $(1 - p_j^{FN}) p_{ij}^{l \rightarrow k}$. We denote by $p_{ij}^{0 \rightarrow k}$ the probability that the channel condition is $\sigma_{ik}$ at time $t+1$, knowing that $y_j^t = 0$ and $y_j^{t+1} = 1$. Then, the state transition from $w_{ij}^t = (0, -\infty)$ to $w_{ij}^{t+1} = (1, \sigma_{ik})$ has probability $p_j^{NF} p_{ij}^{0 \rightarrow k}$. The state transition from $w_{ij}^t = (1, \sigma_{il})$ to $w_{ij}^{t+1} = (0, -\infty)$ has probability $p_j^{FN}$. The transition matrix of the channel state is given by

$$P_{ij}^{ch} = \left[ p_{ij}^{ch}(w_{ij}^{t+1} \mid w_{ij}^t) \right]_{(K+1) \times (K+1)} = \begin{bmatrix} 1 - p_j^{NF} & p_j^{NF} p_{ij}^{0 \rightarrow 1} & p_j^{NF} p_{ij}^{0 \rightarrow 2} & \cdots & p_j^{NF} p_{ij}^{0 \rightarrow K} \\ p_j^{FN} & (1 - p_j^{FN}) p_{ij}^{1 \rightarrow 1} & (1 - p_j^{FN}) p_{ij}^{1 \rightarrow 2} & \cdots & (1 - p_j^{FN}) p_{ij}^{1 \rightarrow K} \\ p_j^{FN} & (1 - p_j^{FN}) p_{ij}^{2 \rightarrow 1} & (1 - p_j^{FN}) p_{ij}^{2 \rightarrow 2} & \cdots & (1 - p_j^{FN}) p_{ij}^{2 \rightarrow K} \\ \vdots & \vdots & \vdots & \ddots & \vdots \\ p_j^{FN} & (1 - p_j^{FN}) p_{ij}^{K \rightarrow 1} & (1 - p_j^{FN}) p_{ij}^{K \rightarrow 2} & \cdots & (1 - p_j^{FN}) p_{ij}^{K \rightarrow K} \end{bmatrix} \quad (3)$$

with $\left[ P_{ij}^{ch} \right]_{1,1}$ being the transition probability from the state $w_{ij}^t = (0, -\infty)$ to $w_{ij}^{t+1} = (0, -\infty)$, $\left[ P_{ij}^{ch} \right]_{1,k}$ being the transition probability from the state $w_{ij}^t = (0, -\infty)$ to $w_{ij}^{t+1} = (1, \sigma_{ik})$, $\left[ P_{ij}^{ch} \right]_{l,1}$ being the transition probability from the state $w_{ij}^t = (1, \sigma_{il})$ to $w_{ij}^{t+1} = (0, -\infty)$, and $\left[ P_{ij}^{ch} \right]_{l,k}$ being the transition probability from the state $w_{ij}^t = (1, \sigma_{il})$ to $w_{ij}^{t+1} = (1, \sigma_{ik})$. We also have $\sum_{k=1}^{K} p_{ij}^{0 \rightarrow k} = 1$ and $\sum_{k=1}^{K} p_{ij}^{l \rightarrow k} = 1$. The randomized version of the channel state profile is denoted by $\boldsymbol{W}_i^t = \left[ W_{i1}^t, ..., W_{iN}^t \right]$. By assuming that the channel evolves independently, the probability of the channel state profile transition is expressed as:

$$p_i^{ch}(\boldsymbol{w}_i^{t+1} \mid \boldsymbol{w}_i^t) = \prod_{j=1}^{N} p_{ij}^{ch}(w_{ij}^{t+1} \mid w_{ij}^t). \tag{4}$$

*(2) Model for packet arrival*

In this subsection, we discuss the modeling of the packet arrival, which is related to the buffer state $v_i^t$. We assume that the data packets of SU $i$ randomly arrive and the arrival times are modeled as a Poisson process with the average arrival rate $\mu_i$ packets/second [16]. However, note that the Poisson process is simply used for illustration purposes and other traffic models (e.g. renewal process, etc.) can also be used in our framework. The number of packets arriving during one time slot is a random variable independent of the time $t$ and denoted as $A_i^t$. The distribution of $A_i^t$ is easily computed as $F_{A_i^t}(n) = (\mu_i \Delta T)^n \exp(\mu_i \Delta T) / n!$, which is independent of $t$. The average number of packets arriving during one time slot equals $\mu_i \Delta T$ [16].

*(3) State transition*

We will now discuss the state transition process. Remember that the state of SU $i$ includes the buffer state $v_i^t$ and the channel state $\boldsymbol{w}_i^t$. In this paper, we assume that the channel state transition is independent of the buffer state transition. In the above, we describe the transition of the channel state $\boldsymbol{w}_i^t$. The buffer state transition is determined by the number of packets arriving and the channel allocation $\boldsymbol{z}_i^t$ during that time slot. The evolution of the buffer state is captured by the equation $v_i^{t+1} = \min\{(v_i^t - R_i(c_{i\delta(\boldsymbol{z}_i^t)}^t)\Delta T)^+ + A_i^t, B_i\}$, where $\delta(\boldsymbol{z}_i^t)$ is the mapping from the channel allocation $\boldsymbol{z}_i^t$ to the index of the specific channel that is allocated to SU $i$ and $\delta(\boldsymbol{0}) = 0$; $R_i(c_{ij}^t)$ is the transmission rate (packets/second) which can be computed as in [31] given the channel SNR $c_{ij}^t$ and $R_i(-\infty) = 0$. We define $h = v_i^{t+1} - (v_i^t - R_i(c_{i\delta(\boldsymbol{z}_i^t)}^t)\Delta T)^+$. Based on the packet arrival model, the buffer state transition probability is computed as

$$p_i^{buf}(v_i^{t+1} \mid v_i^t, \boldsymbol{z}_i^t) = \begin{cases} \dfrac{(\mu_i \Delta T)^h e^{-\mu_i \Delta T}}{h!}, & if\ 0 \le h < B_i - (v_i^t - R_i(c_{i\delta(\boldsymbol{z}_i^t)}^t)\Delta T)^+ \\ \displaystyle\sum_{k=h}^{\infty} \dfrac{(\mu_i \Delta T)^k e^{-\mu_i \Delta T}}{k!}, & if\ h = B_i - (v_i^t - R_i(c_{i\delta(\boldsymbol{z}_i^t)}^t)\Delta T)^+ \end{cases}. \tag{5}$$

Finally, the state evolution of SU $i$ can be expressed as

$$\begin{bmatrix} v_i^{t+1} \\ \boldsymbol{w}_i^{t+1} \end{bmatrix} = \begin{bmatrix} \min\{(v_i^t - R_i(c_{i\delta(\boldsymbol{z}_i^t)}^t)\Delta T)^+ + A_i^t, B_i\} \\ \boldsymbol{W}_i^{t+1}(\boldsymbol{w}_i^t) \end{bmatrix}. \tag{6}$$

The first row in the equation above represents the evolution of the buffer state and the second row represents the channel state evolution. From this expression, we clearly note that the state evolution of the SU depends on both the environment disturbance (i.e. $A_i$ and $\boldsymbol{W}_i^{t+1}$) and other SUs' states and bidding policies through the channel allocation $z_i^t$.

The state transition can be computed as

$$q_i(s_i^{t+1} \mid s_i^t, z_i^t) = \underbrace{p_i^{buf}(v_i^{t+1} \mid v_i^t, z_i^t)}_{\text{buffer state}} \underbrace{p_i^{ch}(w_i^{t+1} \mid w_i^t)}_{\text{channel state}}, \tag{7}$$

where the first term represents the buffer state transition, which is independent of the second term of the channel state transition.

### C. Stage Reward

By playing the auction game in the current stage, SU $i$ receives the channel allocation $z_i^t$ and needs to pay the corresponding payment $\left| \tau_i^t \right|$. Based on the allocated time, the SU transmits the available packets in the buffer. In the next time slot, new packets arrive into the buffer. Newly incoming packets may lead to packets already existing in the buffer being dropped when the buffer is full or their delay deadline has passed. The number of packets lost is denoted by $n_i^t$. Clearly, the performance of the application (e.g. video quality) improves when fewer packets are lost. Hence, we can interpret a negative value of the number of lost packets as the stage gain, which is denoted by $g_i^t$, i.e $g_i^t = -n_i^t$. The reward at time $t$ for SU $i$ is expressed using the quasi-linear form $r_i^t = g_i^t + \tau_i^t$. Note that the gain $g_i^t$ and payment $\tau_i^t$ depend on the states and bids of all the competing SUs in the cognitive radio network. Recall that $\tau_i^t$ is non-positive, we may also interpret $-(g_i^t + \tau_i^t) = n_i^t - \tau_i^t$ as the cost incurred at stage $t$.

Specifically, in state $s_i^t$, SU $i$ knows the number of packets to be transmitted, which are present in the buffer, $v_i^t$, and the channel state. We then compute the stage reward $r_i^t$ by playing the one stage auction game described in Section II.A. Recall that SU $i$ receives the channel allocation $z_i^t$ and payment $\tau_i^t$ from the auction game. The number of packets lost is computed as

$$n_i^t(s_i^t, z_i^t) = \max\{(v_i^t - R_i(c_{i\delta(z_i^t)}^t)\Delta T)^+ + A_i^{t+1} - B_i, 0\}, \tag{8}$$

The gain $g_i^t$ is the negative value of the number of packets lost. The stage reward is given by:

$$r_i = g_i^t(s_i^t, z_i^t) + \tau_i^t = -n_i^t(s_i^t, z_i^t) + \tau_i^t. \tag{9}$$

### D. Selection of Bidding Policy

In the cognitive radio network, we assume that the stochastic game is played by all SUs for an infinite number of stages. This assumption is reasonable for applications having a long duration, such as video streaming. In our network setting, we define a history of the stochastic game up to time $t$ as $h^t = \{s^0, b^0, z^0, \tau^0, ..., s^{t-1}, b^{t-1}, z^{t-1}, \tau^{t-1}, s^t\} \in \mathcal{H}^t$, which summarizes all previous states and the actions taken by the SUs as well as the outcomes at each stage of the auction game and $\mathscr{H}^t$ is the set of all possible history up to time $t$. However, during the stochastic game, each SU $i$ cannot observe the entire history, but rather part of the history $h^t$. The observation of SU $i$ is denoted as $o_i^t \in \mathscr{O}_i^t$ and $o_i^t \subset h^t$. Note that the current state $s_i^t$ can be always observed, i.e. $s_i^t \in o_i^t$. Then, a bidding policy $\pi_i^t : \mathscr{O}_i^t \mapsto \mathscr{B}_i$ for SU $i$ at the time $t$ is defined as a mapping from the observations up to the time $t$ into the specific action, i.e. $b_i^t = \pi_i^t(o_i^t)$. Furthermore, a policy profile $\pi_i$ for SU $i$ aggregates the bidding policies about how to play the game over the entire course of the stochastic game, i.e. $\pi_i = (\pi_i^0, ..., \pi_i^t, ...)$. The policy profile for all the SUs at time slot $t$ is denoted as $\pi^t = (\pi_1^t, ..., \pi_M^t) = (\pi_i^t, \pi_{-i}^t)$.

The policy $\pi_i$ is said to be Markov if the bidding policy $\pi_i^t$ for $\forall t$ is, given the current state $s_i^t$, independent of the states and actions prior to the time $t$, i.e. $\pi_i^t(o_i^t) = \pi_i^t(s_i^t)$. The reward for SU $i$ at the time slot $t$ is $r_i((s_i^t, s_{-i}^t), (b_i^t, b_{-i}^t))$, where the superscript $t$ denotes the time. This reward $r_i((s_i^k, s_{-i}^k), (b_i^k, b_{-i}^k))$ of the stage $k$ is discounted by factor $(\alpha_i)^{k-t}$ at time $t$. The factor $\alpha_i \, (0 \leq \alpha_i < 1)$ is the discounted factor determined by a specific application (for instance, for video streaming applications, this factor can be set based on the tolerable delay). The total discounted sum of rewards $Q_i^t(s^t, (\pi_i^t, \pi_{-i}^t))$ for SU $i$ can be calculated at time $t$ from the state profile $s^t$ as:

$$
\begin{aligned}
Q_i^t(s^t, (\pi_i^t, \pi_{-i}^t)) &= \sum_{k=t}^{\infty} (\alpha_i)^{k-t} r_i((s_i^k, s_{-i}^k), (\pi_i^t(s_i^k), \pi_{-i}^t(s_{-i}^k))) \\
&= \{\sum_{k=t}^{\infty} (\alpha_i)^{k-t} [g_i(s_i^k, z_i^k(\pi^t(s^k))) + \tau_i^t(\pi^t(s^k))]\} \\
&= \underbrace{\{g_i(s_i^t, z_i^t(\pi^t(s^k))) + \tau_i^t(\pi^t(s^k))}_{\text{stage reward at time t}} \\
&\quad + \alpha_i \underbrace{\sum_{s^{t+1} \in \mathscr{S}} \prod_{k=1}^{M} q_k(s_k^{t+1} \mid s_k^t, z_k^t(\pi^t(s^t))) Q_i^{t+1}(s^{t+1}, \pi^t)\}}_{\text{expected future reward}}, \forall s^t \in \mathscr{S},
\end{aligned}
\tag{10}
$$

where $\pi_{-i}^t(s_{-i}^t) = (\pi_1^t(s_1^t), ..., \pi_{i-1}^t(s_{i-1}^t), \pi_{i+1}^t(s_{i+1}^t) ..., \pi_M^t(s_M^t))$. We assume that the SUs implement the policy $\pi^t$ in the subsequent time slots. The total discounted sum of rewards in Eq. (10) consists of two parts: (i) the current stage reward and (ii) the expected future reward discounted by $\alpha_i$. Note that SU $i$ cannot independently determine the above value without explicitly knowing the policies and states of other SUs. The SU maximizes the total discounted sum of future rewards in order to select the bidding policy, which explicitly considers the impact of the current bid vector on the expected future rewards.

We define the *best response* $\beta_i$ for SU $i$ to other SUs' policies $\pi_{-i}^t$ as

$$
\beta_i(\pi_{-i}^t) = \arg\max_{\pi_i} Q_i^t(s^t, (\pi_i^t, \pi_{-i}^t))
\tag{11}
$$

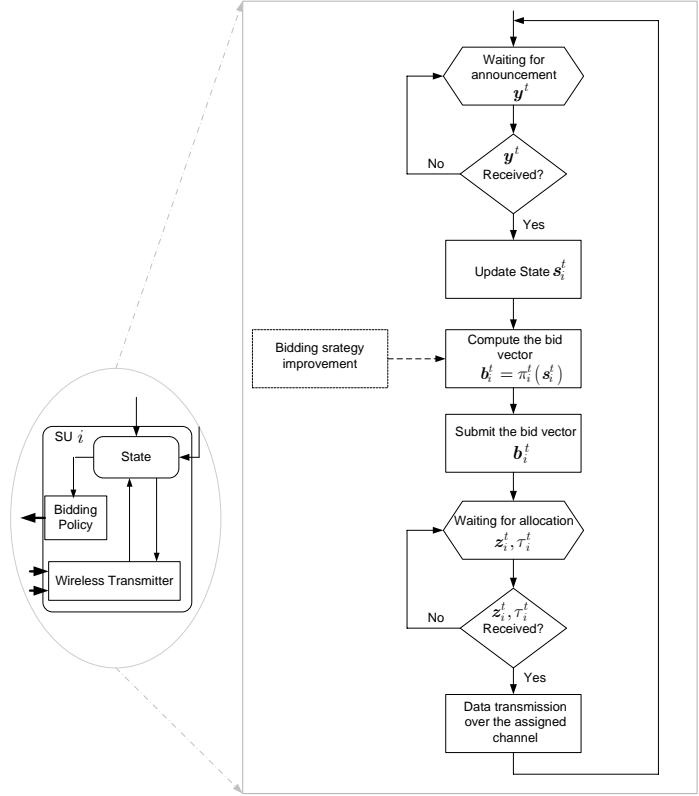

Figure 3. The procedure for SU $i$ to play the auction game at time slot $t$

The central issue in our stochastic game is how the best response policies can be determined by the SUs. In the repeated auction mechanism discussed in Section II.A , the procedure each SU $i$ follows to compete for the channel opportunities is illustrated in Figure 3. In this procedure, the bidding strategy $\pi_i^t$ is continuously improved by the "bidding strategy improvement" module. In Section III.E, we discuss the challenges involved in building such a module, and in Section IV we develop a best response learning algorithm that can be used for improving the bidding strategy.

## E. Challenges for Selecting the Bidding Policy

Recall that during each time slot, the CSM announces an auction based on the available spectrum opportunities and then SUs bid for the resources. To enable the successful deployment of this resource auction mechanism, we can prove, similarly to our prior work in [28], that SUs have no incentive to misrepresent their information, i.e. they adhere to the "truth telling" policy. We assume that at each time slot $t$ , SU $i$ has preference $u_{ij}^t$ over the channel $j$ , which capture the benefit derived when using that channel. The preference $u_{ij}^t$ is interpreted as the benefit obtained by SU $i$ when using channel $j$ , compared to the benefit when this channel is not used. Note that this benefit also includes the expected future rewards. The optimal bid $b_{ij}^{t,opt}$ that SU $i$ can take on the channel $j$ at time $t$ is the bid maximizing the net benefit $u_{ij}^t + \tau_i^t$. In auction discussed in Section II.A, the optimal bid that SU $i$ can make is $b_{ij}^{t,opt} = u_{ij}^t$, i.e. the optimal bid for SU $i$ is to announces its true preference to the CSM [28]. The proof is omitted here due to space limitations, since it is similar to that in [28]. The payment made by SU $i$ is computed by the CSM based on the inconvenience incurred by other SUs due to SU $i$ during that time slot [32].

Next, we define the preference $u_{ij}^t$ in the context of the stochastic game model. Using the channel $j$, SU $i$ obtains the immediate gain $g_i^t(s_i^t, e_j)$ by transmitting the packets in its buffer, where $e_j$ indicates that channel $j$ is allocated to SU $i$ during the current time slot. SU $i$ then moves into next state $s_i^{t+1}$ from which it may obtain the future reward $Q_i^{t+1}((s_i^{t+1}, s_{-i}^{t+1}), \pi^t)$. On the other hand, if no channel is assigned to SU $i$, it receives the immediate gain $g_i^t(s_i^t, \mathbf{0})$ and then moves into the next state $s_i^{t+1}$ from which it may obtain the future reward $Q_i^{t+1}((s_i^{t+1}, s_{-i}^{t+1}), \pi^t)$. We define a feasible set of channel assignments to SU $i$'s opponents, given SU $i$'s channel allocation $(z_i^t)$, as $\mathcal{Z}_{-i}^t(z_i^t)$, with $\mathcal{Z}_{-i}^t(z_i^t) = \{Z_{-i}^t \mid \sum_{k=1, k \neq i}^{M} z_{kj}^t = y_j^t - z_i^t, \forall j, \sum_{j=1}^{N} z_{kj}^t \leq 1, \forall k \neq i, z_{kj}^t \in \{0,1\}\}$.

The preference over the current state can be then computed as

$$u_{ij}^t(s_i^t, s_{-i}^t) = \left[ g_i(s_i^t, e_j) + \alpha_i \sum_{s^{t+1} \in \mathcal{S}} q_i(s_i^{t+1} \mid s_i^t, e_j) \sum_{Z_{-i}^t \in \mathcal{Z}_{-i}^t(e_j)} \prod_{k=1}^{M} q_k(s_k^{t+1} \mid s_k^t, z_k^t) Q_i^{t+1}((s_i^{t+1}, s_{-i}^{t+1}), \pi^t) \right]$$
$$- \left[ g_i(s_i^t, \mathbf{0}) + \alpha_i \sum_{s^{t+1} \in \mathcal{S}} q_i(s_i^{t+1} \mid s_i^t, \mathbf{0}) \sum_{Z_{-i}^t \in \mathcal{Z}_{-i}^t(\mathbf{0})} \prod_{k=1}^{M} q_k(s_k^{t+1} \mid s_k^t, z_k^t) Q_i^{t+1}((s_i^{t+1}, s_{-i}^{t+1}), \pi^t) \right]. \quad (12)$$

From this equation, it is clear that the true value $u_{ij}^t$ depends on its own current state $s_i^t$, but also the other SUs' states $s_{-i}^t$, the channel allocations $\mathcal{Z}_{-i}^t(e_j)$ to the other users when channel $j$ is assigned to SU $i$, $\mathcal{Z}_{-i}^t(\mathbf{0})$ when SU $i$ is not assigned to any channel, and the state transition models $q_k(s_k^{t+1} \mid s_k^t, z_k^t), \forall k$. However, the other SUs' states, the channel allocations and the state transition models of other SUs are not known to SU $i$, and it is thus impossible for each SU to determine its preference $u_{ij}^t(s_i^t, s_{-i}^t)$.

The optimal bid on channel $j$ is $b_{ij}^{t,opt} = u_{ij}^t(s_i^t, s_{-i}^t)$. Without knowing the other SUs' states and state transition models, SU $i$ cannot derive its optimal bidding strategy. However, if SU $i$ chooses the bid vector by only maximizing the immediate reward $g_i^t + \tau_i^t$, i.e. the total discounted sum of reward degenerates in $Q_i^t(s^t, \pi^t) = g_i(s_i^t, z_i^t(\pi^t(s^t))) + \tau_i^t(\pi^t(s^t))$ by setting $\alpha_i = 0$. Then, the preference over channel $j$ becomes $u_{ij}^t(s_i^t, s_{-i}^t) = g_i(s_i^t, e_j) - g_i(s_i^t, \mathbf{0})$. Since now $u_{ij}^t$ only depends on the state $s_i^t$, SU $i$ can compute both the optimal bid vector as well as the optimal bidding policy. We refer to this optimal bidding policy as the "myopic" policy, since it only takes the immediate reward into consideration and ignores the future impact. The myopic policy is referred to as $\pi_i^{myopic}$. To solve the difficult problem of optimal bidding policy selection when $\alpha_i \neq 0$, an SU needs to forecast the impact of its current bidding actions on the expected future rewards discounted by $\alpha_i$. The forecast can be performed using learning from its past experiences.

## IV. INTERACTIVE LEARNING FOR PLAYING THE RESOURCE MANAGEMENT GAME

### A. How to Evaluate Learning Algorithms?

In Section III.E, it was shown that an SU needs to know other SUs' states and state transition models in order to derive its own optimal bidding policy. This coupling among SUs is due to the shared nature of the wireless resources. However, an SU cannot exactly know the other SUs' models and private information in the wireless networks. Thus, to improve the bidding policy, an SU can only predict the impacts of dynamics (uncertainties) caused by the competing SUs based on its observations from past auctions. In this paper, we propose a learning algorithm for predicting these uncertainties. We define a learning algorithm $\mathcal{L}_i$ for SU $i$ as a function taking the observation $o_i^t$ as input and having the bidding policy $\pi_i^t$ as output.

Before developing a learning algorithm, we first discuss how to evaluate the performance of a learning algorithm in terms of its impact on the SU's reward. Unlike existing multi-agent learning research, which is aimed at achieving converge to an equilibrium point for the interacting agents, we develop learning algorithms based on the performance of the bidding strategy on the SU's reward. We denote a bidding policy generated by the learning algorithm $\mathcal{L}_i$ as $\pi_i^{\mathcal{L}_i}$. An SU will learn in order to improve its bidding policy and its rewards from participating in the auction game. The performance of the bidding strategy $\pi_i$ is defined as the time average reward that SU $i$ obtains in a time window with length $T$ when it adopts $\pi_i$:

$$\mathcal{V}^{\pi_i}(T) = \frac{1}{T}\sum_{k=1}^{T} r_i^k \tag{13}$$

Using this definition, the performance of two learning algorithms can be easily compared. For instance, given two algorithm $\mathcal{L}_i'$ and $\mathcal{L}_i''$, if $\mathcal{V}^{\pi_i^{\mathcal{L}_i'}} > \mathcal{V}^{\pi_i^{\mathcal{L}_i''}}$, then we say that learning algorithm $\mathcal{L}_i'$ is better than $\mathcal{L}_i''$.

### B. What Information to Learn from?

First let us consider what information the SU can observe while playing the stochastic game in our cognitive radio network. As shown in Figure 3, at the beginning of time slot $t$, the SUs submit the bids $b_i^t, \forall i$. Then, the CSM returns the channel allocation $z_i^t, \forall i$ and $\tau_i^t, \forall i$. If SU $i$ is not allowed to observe the bids, the channel allocations and payments for other SUs, then the observation of SU $i$ becomes $\boldsymbol{o}_i^t = \{\boldsymbol{s}_i^0, \boldsymbol{b}_i^0, \boldsymbol{z}_i^0, \boldsymbol{\tau}_i^0, ..., \boldsymbol{s}_i^{t-1}, \boldsymbol{b}_i^{t-1}, \boldsymbol{z}_i^{t-1}, \boldsymbol{\tau}_i^{t-1}, \boldsymbol{s}_i^t\}$. If the information is exchanged among SUs or broadcasted and overheard by all SUs, the observed information by SU $i$ becomes $\boldsymbol{o}_i^t = \{\boldsymbol{s}^0, \boldsymbol{b}^0, \boldsymbol{z}^0, \boldsymbol{\tau}^0, ..., \boldsymbol{s}^{t-1}, \boldsymbol{b}^{t-1}, \boldsymbol{z}^{t-1}, \boldsymbol{\tau}^{t-1}, \boldsymbol{s}_i^t\}$. Now, the problem that needs to be solved by SU $i$ is how it can improve its own policy for playing the game by learning from the observation $\boldsymbol{o}_i^t$. In this paper, we assume that SU $i$ observes the information $\boldsymbol{o}_i^t = \{\boldsymbol{s}_i^0, \boldsymbol{b}_i^0, \boldsymbol{z}_i^0, \boldsymbol{\tau}_i^0, ..., \boldsymbol{s}_i^{t-1}, \boldsymbol{b}_i^{t-1}, \boldsymbol{z}_i^{t-1}, \boldsymbol{\tau}_i^{t-1}, \boldsymbol{s}_i^t\}$.

## C. What to Learn?

In Section IV.A, we introduced learning as a tool to predict the impacts of dynamics and hence, improve the bidding policy. However, a key question is what needs to be learned. Recall that the optimal bidding policy for SU $i$ is to generate a bid vector that represents its preferences for using different channels. From Eq. (12), we can see that SU $i$ needs to learn: (i) the state space of other SUs, i.e. $\boldsymbol{\mathcal{S}}_{-i}$; (ii) the current state of other SUs, i.e. $\boldsymbol{s}_{-i}^t$; (iii) the transition probability of other SUs, i.e. $\prod_{k \neq i} q_k \left( \boldsymbol{s}_k^{t+1} \mid \boldsymbol{s}_k^t, \boldsymbol{z}_k^t \right)$; (iv) the resource allocation $\boldsymbol{\mathcal{Z}}_{-i}^t(\boldsymbol{e}_j), \forall j$ and $\boldsymbol{\mathcal{Z}}_{-i}^t(\boldsymbol{0})$; and (v) the discounted sum of rewards $Q_i^{t+1} \left( \left( \boldsymbol{s}_i^{t+1}, \boldsymbol{s}_{-i}^{t+1} \right), \boldsymbol{\pi}^t \right)$.

However, SU $i$ can only observes the information $\boldsymbol{o}_i^t = \{\boldsymbol{s}_i^0, \boldsymbol{b}_i^0, \boldsymbol{z}_i^0, \boldsymbol{\tau}_i^0, ..., \boldsymbol{s}_i^{t-1}, \boldsymbol{b}_i^{t-1}, \boldsymbol{z}_i^{t-1}, \boldsymbol{\tau}_i^{t-1}, \boldsymbol{s}_i^t\}$ from which SU $i$ cannot accurately infer the other SUs' state space and transition probability. Moreover, capturing the exact information about other SUs requires heavy computational and storage complexity. Instead, we allow SU $i$ to classify the space $\boldsymbol{\mathcal{S}}_{-i}$ into $H_i$ classes each of which is represented by a representative state $\tilde{s}_{-i,h}, h \in \{1, ..., H_i\}$. We discuss how the space $\boldsymbol{\mathcal{S}}_{-i}$ is decomposed in Section IV.D. By dividing the state space $\boldsymbol{\mathcal{S}}_{-i}$, the transition probability $\prod_{k \neq i} q_k \left( \boldsymbol{s}_k^{t+1} \mid \boldsymbol{s}_k^t, \boldsymbol{z}_k^t \right)$ is approximated by $q_{-i} \left( \tilde{s}_{-i}^{t+1} \mid \tilde{s}_{-i}^t, \boldsymbol{z}_i^t \right)$, where $\tilde{s}_{-i}^t$ and $\tilde{s}_{-i}^{t+1}$ are the representative states of the classes that $\boldsymbol{s}_{-i}^t$ and $\boldsymbol{s}_{-i}^{t+1}$ belong to. This approximation is performed by aggregating all other SUs' states into one representative state and assuming that the transition depends on the resource allocation $\boldsymbol{z}_i^t$. The transition probability approximation is also discussed in Section IV.D. The discounted sum of rewards $Q_i^{t+1} \left( \left( \boldsymbol{s}_i^{t+1}, \boldsymbol{s}_{-i}^{t+1} \right), \boldsymbol{\pi}^t \right)$ is approximated by $V_i^{t+1} \left( \left( \boldsymbol{s}_i^{t+1}, \tilde{s}_{-i}^{t+1} \right) \right)$. Note that the classification on the state space $\boldsymbol{\mathcal{S}}_{-i}$ and approximation of the transition probability and discounted sum of rewards affects the learning performance. Hence, a user can tradeoff an increased complexity for an increased performance. After the classification, the preference computation can be approximated as

$$u_{ij}^t\left(s_i^t, \tilde{s}_{-i}^t\right) = \left[g_i\left(s_i^t, e_j\right) + \alpha_i \sum_{\left(s_i^{t+1}, \tilde{s}_{-i}^{t+1}\right) \in \mathcal{S}} q_i\left(s_i^{t+1} \mid s_i^t, e_j\right) q_{-i}\left(\tilde{s}_i^{t+1} \mid \tilde{s}_i^t, e_j\right) V_i^{t+1}\left(s_i^{t+1}, \tilde{s}_{-i}^{t+1}\right)\right]$$
$$- \left[g_i\left(s_i^t, \mathbf{0}\right) + \alpha_i \sum_{\left(s_i^{t+1}, \tilde{s}_{-i}^{t+1}\right) \in \mathcal{S}} q_i\left(s_i^{t+1} \mid s_i^t, \mathbf{0}\right) q_{-i}\left(\tilde{s}_i^{t+1} \mid \tilde{s}_i^t, \mathbf{0}\right) V_i^{t+1}\left(s_i^{t+1}, \tilde{s}_{-i}^{t+1}\right)\right]. \tag{14}$$

In this setting, to find the approximated preference and thus, the approximated optimal bidding policy, we need to learn the following from the past observations: (i) how the space $\mathcal{S}_{-i}$ is classified; (ii) the transition probability $q_{-i}\left(\tilde{s}_{-i}^{t+1} \mid \tilde{s}_{-i}^t, z_i^t\right)$; (iii) the average future rewards $V_i^{t+1}\left(\left(s_i^{t+1}, \tilde{s}_{-i}^{t+1}\right)\right)$.

### D. How to Learn?

In this section, we develop a learning algorithm to estimate the terms listed in Section IV.C.

#### (1) Decomposition of the space $\mathcal{S}_{-i}$

As discussed in Section IV.B, only $o_i^t = \{s_i^0, b_i^0, z_i^0, \tau_i^0, ..., s_i^{t-1}, b_i^{t-1}, z_i^{t-1}, \tau_i^{t-1}, s_i^t\}$ are observed. From the auction mechanism presented in Section II.A, we know that the value of the tax $\tau_i^t$ is computed based on the inconvenience that SU $i$ causes to the other SUs. In other words, a higher value of $\left|\tau_i^t\right|$ indicates that the network is more congested[9]. Based on the bid vector $b_i^t$, the channel allocation $z_i^t$ and the tax $\tau_i^t$, SU $i$ can infer the network congestion and thus, indirectly, the resource requirements of the competing SUs. Instead of knowing the exact state space of other SUs, SU $i$ can classify the space $\mathcal{S}_{-i}$ as follows.

We assume the maximum absolute tax is $\Gamma$. We split the range $[0, \Gamma]$ into $[\Gamma_0, \Gamma_1), [\Gamma_1, \Gamma_2), ..., [\Gamma_{H_i - 1}, \Gamma_{H_i}]$ with $0 = \Gamma_0 \leq \Gamma_1 \leq \cdots \leq \Gamma_{H_i} = \Gamma$. Here, we assume that the values of $\{\Gamma_1, ..., \Gamma_{H_i - 1}\}$ are equally located in the range of $[0, \Gamma]$. (Note that more sophisticated selection for these values can be deployed, and this forms an interesting area of future research.)

We need to consider three cases to determine the representative state $\tilde{s}_{-i}^t$ at time $t$.

(1) If the resource allocation $z_i^t \neq \mathbf{0}$, then the representative state of other SUs is chosen as
$$\tilde{s}_{-i}^t = h, \text{ if } \left|\tau_i^t\right| \in [\Gamma_{h-1}, \Gamma_h). \tag{15}$$

(2) If the resource allocation $z_i^t = \mathbf{0}$ but $y^t \neq \mathbf{0}$, the tax is 0. In this case, we cannot use the tax to predict the network congestion. However, we can infer that the congestion is more severe than the minimum bid for those available channels, i.e. $\min_{j \in \{l: y_l^t \neq 0\}} \{b_{ij}^t\}$. This is because, in this current stage of the auction game, only SU $i'$ with $b_{i'j}^t \geq b_{ij}^t$ can obtain channel $j$ which indicates that $\left|\tau_i^t\right| \geq \min_{j \in \{l: y_l^t \neq 0\}} \{b_{ij}^t\}$, if SU $i$ is allocated any channel. Then the representative state of other SUs is chosen as

---

[9] When the CSM deploys a mechanism without tax for the resource management, the space classification for other SUs can also be done based on the announced information and corresponding resource allocation.

$$\tilde{s}_{-i}^{t} = h, \; if \; \min_{j \in \{l : y_{i}^{l} \neq 0\}} \left\{b_{ij}^{t}\right\} \in [\Gamma_{h-1}, \Gamma_{h}) \tag{16}$$

(3) If the resource allocation $\boldsymbol{z}_{i}^{t} = \boldsymbol{0}$ and $\boldsymbol{y}^{t} = \boldsymbol{0}$, there is no interaction among the SUs in this time slot. Hence, $\tilde{s}_{-i}^{t} = \tilde{s}_{-i}^{t-1}$.

*(2) Estimating the transition probability*

To estimate the transition probability, SU $i$ maintains a table $F$ with size $H_{i} \times H_{i} \times (N+1)$. Each entry $f_{h',h'',j}$ in the table $F$ represents the number of transitions from state $\tilde{s}_{-i}^{t} = h''$ to $\tilde{s}_{-i}^{t+1} = h'$ when the resource allocation $\boldsymbol{z}_{i}^{t} = \boldsymbol{e}_{j}$ (or $\boldsymbol{0}$ if $j = 0$). It is clear that $H_{i}$ will influence significantly the complexity and memory requirements etc. of SU $i$. The update of $F$ is simply based on the observation $\boldsymbol{o}_{i}^{t}$ and the state classification in the above section. Then, we use the frequency to approximate the transition probability [21], i.e.

$$q_{-i}\left(\tilde{s}_{-i}^{t+1} = h' \mid \tilde{s}_{-i}^{t} = h'', \boldsymbol{e}_{j}\right) = \frac{f_{h',h'',j}}{\sum\limits_{h'} f_{h',h'',j}} \tag{17}$$

*(3) Learning the future reward*

By classifying the state space $\mathcal{S}_{-i}$ and estimating the transition probability, SU $i$ can now forecast the value of the average future reward $V_{i}^{t+1}\left(\left(\boldsymbol{s}_{i}^{t+1}, \tilde{s}_{-i}^{t+1}\right)\right)$ using learning. Eq. (10) can be approximated by

$$Q_{i}^{t}\left(\boldsymbol{s}_{i}^{t}, \tilde{s}_{-i}^{t}\right) = \left\{g_{i}(\boldsymbol{s}_{i}^{t}, \boldsymbol{z}_{i}^{t}) + \tau_{i}^{t} + \alpha_{i} \sum_{\left(\boldsymbol{s}_{i}^{t+1}, \tilde{s}_{-i}^{t+1}\right) \in \mathcal{S}} q_{i}\left(\boldsymbol{s}_{i}^{t+1} \mid \boldsymbol{s}_{i}^{t}, \boldsymbol{z}_{i}^{t}\right) q_{-i}\left(\tilde{s}_{-i}^{t+1} \mid \tilde{s}_{-i}^{t}, \boldsymbol{z}_{i}^{t}\right) V_{i}^{t+1}\left(\left(\boldsymbol{s}_{i}^{t+1}, \tilde{s}_{-i}^{t+1}\right)\right)\right\} \tag{18}$$

Similar to the Q-learning established in [23], we also use the received rewards to update the estimation of future rewards. However, the main difference between our proposed algorithm and Q-learning is that our solution explicitly considers the impacts of other SUs' bidding actions through the state classifications and transition probability approximation.

We use a 2-dimensional table to store the value $V_{i}\left(\boldsymbol{s}_{i}, \tilde{s}_{-i}\right)$ with $\boldsymbol{s}_{i} \in \mathcal{S}_{i}$, $\tilde{s}_{-i} \in \tilde{S}_{-i}$, where $\tilde{S}_{-i}$ is the set of representative state for other SUs. The total number of entries in $V_{i}$ is $|\mathcal{S}_{i}| \times \left|\tilde{\mathcal{S}}_{-i}\right|$. SU $i$ updates the value of $V_{i}\left(\left(\boldsymbol{s}_{i}, \tilde{s}_{-i}\right)\right)$ at time $t$ according to the following rules:

$$V_{i}^{t}\left(\boldsymbol{s}_{i}, \tilde{s}_{-i}\right) = \begin{cases} \left(1 - \gamma_{i}^{t}\right) V_{i}^{t-1}\left(\boldsymbol{s}_{i}, \tilde{s}_{-i}\right) + \gamma_{i}^{t} Q_{i}^{t}\left(\boldsymbol{s}_{i}^{t}, \tilde{s}_{-i}^{t}\right) & if \left(\boldsymbol{s}_{i}^{t}, \tilde{\boldsymbol{s}}_{-i}^{t}\right) = \left(\boldsymbol{s}_{i}, \tilde{\boldsymbol{s}}_{-i}\right) \\ V_{i}^{t}\left(\boldsymbol{s}_{i}, \tilde{s}_{-i}\right) & otherwise \end{cases} \tag{19}$$

where $\gamma_{i}^{t} \in [0,1)$ is a learning rate factor satisfying $\sum_{t=1}^{\infty} \gamma_{i}^{t} = \infty$ and $\sum_{t=1}^{\infty} \left(\gamma_{i}^{t}\right)^{2} < \infty$ [23], In summary, the learning procedure that is developed for an SU is shown in Table 1.

Table 1. Learning Procedure

---

**Initializing**: $V_i^0\left(\left(\boldsymbol{s}_i, \tilde{s}_{-i}\right)\right) \Leftarrow 0$ for all possible states $\boldsymbol{s}_i \in \mathcal{S}_i$, $\tilde{\boldsymbol{s}}_{-i} \in \tilde{S}_{-i}$.

**Learning**:

At time $t$, SU $i$:

    a.   Observes the current state $\boldsymbol{s}_i^t$;

    b.   Chooses an action $\boldsymbol{b}_i^t = \left[u_{i1}^t, \ldots, u_{iN}^t\right]$ as computed in Eq. (14) by replacing $V_i\left(\boldsymbol{s}_i^{t+1}, \tilde{s}_{-i}^{t+1}\right)$ with $V_i^{t-1}\left(\boldsymbol{s}_i^{t+1}, \tilde{s}_{-i}^{t+1}\right)$, and then submits it to the CSM;

    c.   Receives the allocation $\boldsymbol{z}_i^t$ and payment $\tau_i^t$;

    d.   Computes the representative state $\tilde{s}_{-i}^t$ as in Section IV.D(1) and update the transition probability as in Section IV.D(2);

    e.   Computes the expected total discounted sum of the rewards $Q_i^t\left(\boldsymbol{s}_i^t, \tilde{s}_{-i}^t\right)$ as in Eq. (18);

    f.   Updates the future reward table $V_i\left(\left(\boldsymbol{s}_i, \tilde{s}_{-i}\right)\right)$ at the state $\left(\boldsymbol{s}_i^t, \tilde{s}_{-i}^t\right)$ using the learning rate factor $\gamma_i^t$, according to Eq. (19).

---

# V. Simulation Results

In this section, we aim at quantifying the performance of our proposed stochastic interaction and learning framework. We assume that the SUs compete for the available spectrum opportunities in order to transmit delay-sensitive multimedia data. First, we compare the performance of various bidding strategies. Next, we quantify the performance of our proposed learning algorithm in various network environments. We will present here only several illustrative examples. However, the same observations can be obtained using a larger number of SUs or channels.

## A. Various bidding strategies for dynamic multi-user interaction

In this section, we highlight the merits of the stochastic interaction framework proposed in Section III by comparing the performance of different SUs, which deploy different bidding strategies. In the repeated auction games presented in Section II, the SUs are required to submit the bid vector on the available channels. The SUs can deploy different bidding strategies to generate their bid vector:

1. Fixed bidding strategy $\pi_i^{fixed}$: this strategy generates a constant bid vector during each stage of the auction game, irrespective of the state that SU $i$ is currently in and of the states other SUs are in. In other words, $\pi_i^{fixed}$ does not consider any of the dynamics defined in Section I.

2. Source-aware bidding strategy $\pi_i^{source}$: this strategy generates various bid vectors by considering the

dynamics in source characteristics (based on the current buffer state), but not the channel dynamics.

3. Myopic bidding strategy $\pi_i^{myopic}$ : this strategy takes into account the disturbance due to the environment as well as the impact caused by other SUs, as discussed in Section III.E. However, it does not consider the impact on the future rewards.

4. Bidding strategy based on best response learning $\pi_i^{\mathcal{L}_i}$ : This strategy is produced using the learning algorithm proposed in Section IV. $\pi_i^{\mathcal{L}_i}$ considers the two types of dynamics defined in Section I, and the interaction impact on the future reward.

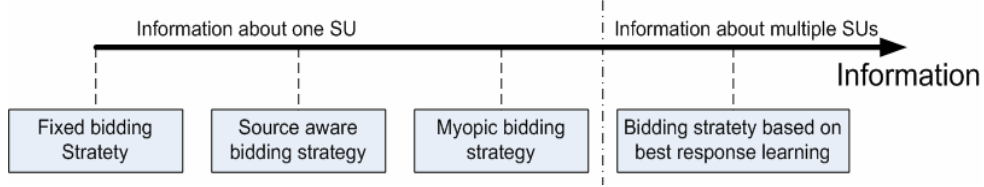

Figure 4. The illustration of bidding strategies based on the required information

In terms of the required information, the above bidding strategies are illustrated in Figure 4. For instance, the fixed bidding strategy $\pi_i^{fixed}$ does not require information about SU $i$'s state or other SUs' states. The source-aware bidding strategy $\pi_i^{buff}$ considers the source characteristics based on the current buffer state. However, the myopic bidding strategy $\pi_i^{myopic}$ requires full information about SU $i$'s state. The bidding strategy based on best response learning $\pi_i^{\mathcal{L}_i}$ also requires information about the states of other SUs.

In this simulation, we consider the cognitive radio network as an extension of WLANs with spectral agile capability [14]. In the following, we first simulate the case that two SUs compete for the channel opportunities and then extend to the case with multiple (five) SUs.

*(1) Competition among two SUs for channel opportunities*

We first consider a simple illustrative network with two SUs competing for the available channel opportunities. The packet arrivals of the SUs are modeled using a Poisson process with the same average arrival rate of 2Mbps. For illustration simplicity, the channel condition of SU 1 (SU 2) on each channel takes only three values ( $K = 3$ ), which are 18dB, 23dB and 26dB. The transition probabilities are $p_{ij}^{0 \rightarrow 1} = p_{ij}^{0 \rightarrow 2} = 0.4, p_{ij}^{0 \rightarrow 3} = 0.2$ , $p_{1j}^{l \rightarrow 1} = p_{1j}^{l \rightarrow 2} = 0.4, p_{1j}^{l \rightarrow 3} = 0.2$ , $\forall i, j, l$ . The transition probability of the availability of the channels to the SUs are $p_j^{NF} = p_j^{FN} = 0.5$ . For illustration simplicity, the environment parameters experienced by the two SUs are the same. The length of the time slot $\Delta T$ is $10^{-2}$ s.

In this simulation, we consider five scenarios. In scenario (1), both SU 1 and 2 deploy the fixed bidding strategy $\pi_1^{fixed}$ . In scenario (2)~(5), SU 1 deploys the fixed bidding strategy $\pi_1^{fixed}$ , source-aware bidding strategy $\pi_1^{source}$ , myopic bidding strategy $\pi_1^{myopic}$ and best response learning based bidding strategy $\pi_1^{\mathcal{L}_i}$ ,

respectively, and SU 2 always deploys the myopic bidding strategy $\pi_2^{myopic}$. The discounted factor for the best response learning algorithm is set to 0.8. As discussed in Section III.C, the stage reward is defined as $r_i^t = (g_i^t + \tau_i^t) = -(n_i^t - \tau_i^t)$, with $(n_i^t - \tau_i^t)$ being the number of packet lost plus the tax charged by the CSM (note that $\tau_i^t \leq 0$). This can be interpreted as the cost incurred at each stage. Similar to Eq. (13), we use the average cost over the time window $T = 1000$ to evaluate the performance of the bidding strategies. Hence, the lower the average cost, the better the performance of the bidding strategy is. The packet loss rate, average tax and cost per time slot are presented in Table 2. The accumulated packet loss and cost of SU 1 for the five scenarios are plotted in Figure 5(a) and (b), respectively.

From this simulation, comparing scenario 2 with scenario 1, we observe that when SU 2 deploys the myopic strategy against SU 1 which adopted the fixed bidding strategy, SU 2 reduces its average cost by around 42% and the average packet loss rate by around 16.6%. This significant improvement is because SU 2 can value the channel opportunities more accurately by modeling and considering its experienced dynamics, i.e. source characteristics, channel conditions and availability.

In scenario 3, SU 1 improves its bidding strategy (i.e. it deploys now a source-aware bidding strategy) by partially considering its experienced environment, i.e. SU 1 generates its bid vector by only considering the source dynamics though its current buffer state. Compared to scenario 2, if SU 1 considers more information about its own state, it can further reduce its packet loss rate by an average of 4.5% and an average cost by around 5.4%. This observation verifies that the information about the SU's state improves the bidding strategy.

In scenario 4, SU 1 deploys a myopic bidding strategy which is more advanced than the source-aware bidding strategy since it considers both types of dynamics defined in Section I (including the dynamics regarding to the source characteristics, channel conditions, and channel availability, and the interaction with other SUs in the auction mechanism). The significant improvement in terms of packet loss rate (13% reduced) and average cost (25% reduced), compared to scenario 2, indicates that the myopic bidding strategy provides the optimal bid vector when only current benefits are considered as shown in Section III.E.

In scenario 5, SU 1 improves further the bidding strategy using the best response learning algorithm developed in Section IV. Using learning, SU 1 reduces the packet loss rate to 15.14% and the average cost to 1.7428 (11.8% lower compared to scenario 4). This significant improvement is due to the ability of the SU to learning and forecast the future impact of its current actions.

It is also worth to note that the reduction of the packet loss rate of SU 1 in scenarios 2~5 comes from two parts: one is the advanced bidding strategies, which allows the SU to take into consideration more information about its own states and the other SUs' states and, based on this, better forecast the impact of various actions, and the other one is the increase in the amount of resources consumed by SU 1 which corresponds to higher tax charged by the CSM, as shown in Table 2.

Table 2. Performance of SU 1 and 2 with various bidding strategies in the two SUs network

|  | Bidding Strategies | SU 1 | | | SU 2 | | |
|---|---|---|---|---|---|---|---|
|  |  | Packet loss rate (%) | Average tax | Average cost | Packet loss rate (10%) | Average tax | Average cost |
| Scenario 1 | $\pi_1^{fixed}, \pi_2^{fixed}$ | 32.53 | 0.4875 | 2.8966 | 31.05 | 0.5095 | 2.6104 |
| Scenario 2 | $\pi_1^{fixed}, \pi_2^{myopic}$ | 34.36 | 0.1222 | 2.6337 | 14.39 | 0.5495 | 1.5105 |
| Scenario 3 | $\pi_1^{source}, \pi_2^{myopic}$ | 29.83 | 0.3147 | 2.4915 | 18.11 | 0.6048 | 1.6116 |
| Scenario 4 | $\pi_1^{myopic}, \pi_2^{myopic}$ | 21.55 | 0.4669 | 1.9767 | 19.55 | 0.3763 | 1.7837 |
| Scenario 5 | $\pi_1^{\mathcal{L}}, \pi_2^{myopic}$ | 15.14 | 0.6923 | 1.7428 | 27.29 | 0.4197 | 2.2967 |

We further note that the bidding strategy deployed by SU 1 will affect the performance of SU 2. For example, comparing scenario 2 with scenario 4, the fixed bidding strategy of SU 1 in scenario 2 leads to a lower average cost (15% reduced) for SU 2. This is because SU 1 uses a fixed bidding strategy, which does not account for the dynamic changes in its environment, while SU 2 minimizes its current cost (the number of packets lost plus the tax) based on its current state. However, when comparing scenario 5 with scenario 4, SU 1 using learning not only improves its prediction of the current environment dynamics but also better predicts the impact on the future cost based on the observations. The improvement leads to higher resource allocation (hence, incurring higher tax, see in Table 2) for SU 1, thereby resulting in worse performance for SU 2 (i.e. the average cost is increased by 22.2%).

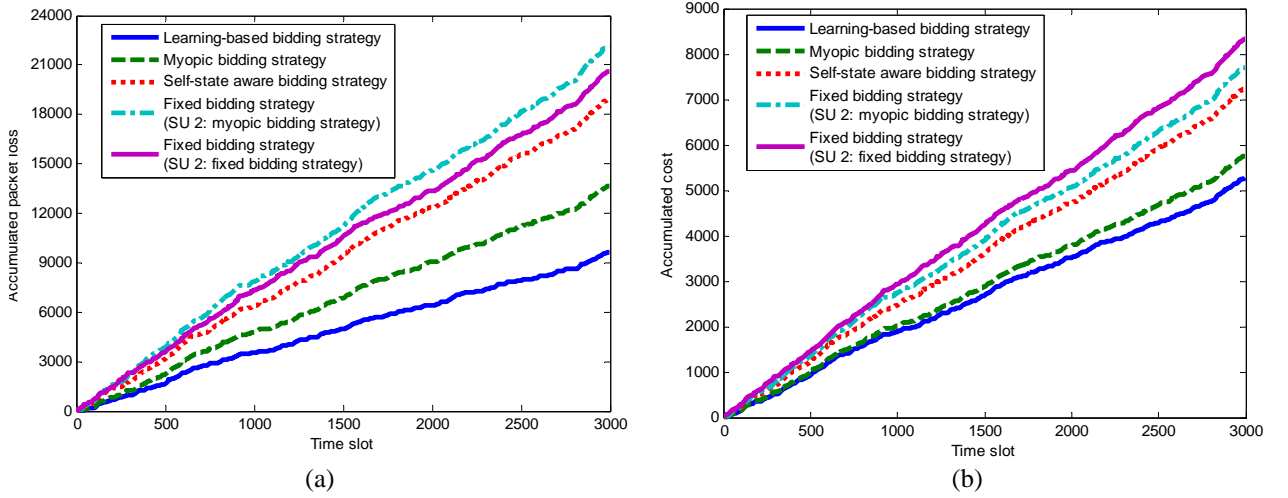

Figure 5. The accumulated packet loss and cost of SU 1 in the five scenarios, (a) accumulated packet loss over the time slot; (b) accumulated cost over the time slot

*(2) Multiple SUs competition for channel opportunities*

In this simulation, we consider five SUs competing for the available channel opportunities in the WLAN-like cognitive radio network. The packet arrivals of all the five SUs are modeled using a Poisson process with the same average arrival rate of 2Mbps. The number of channels is 3 and the channel condition of all the five SUs on each channel takes only three values ($K = 3$), which are 18dB, 23dB and 26dB. The transition probabilities are $p_{ij}^{0 \rightarrow 1} = p_{ij}^{0 \rightarrow 2} = 0.4, p_{ij}^{0 \rightarrow 3} = 0.2$, $p_{1j}^{l \rightarrow 1} = p_{1j}^{l \rightarrow 2} = 0.4, p_{1j}^{l \rightarrow 3} = 0.2$, $\forall i, j, l$. The parameters of the model of the availability of the channels to the SUs are $p_j^{NF} = 0.7, p_j^{FN} = 0.3$. The length of the time slot $\Delta T$ is also $10^{-2}$ s. Similar parameters are used for the five SUs in order to clearly illustrate the performance differences obtained based on the different strategies.

In this simulation, we consider only two scenarios. In scenario (1), all SUs deploy a myopic bidding strategy $\pi_i^{myopic}, i = 1, 2, ..., 5$, while in scenario (2), SU 5 deploys the multi-user learning-based bidding strategy $\pi_5^{\mathcal{L}}$ with the disc and the other SUs deploy the myopic bidding strategy $\pi_i^{myopic}, i = 1, ..., 4$. The packet loss rate and cost per time slot incurred by the SUs are presented in Table 3. The accumulated packet loss and cost of SU 5 for the five scenarios are plotted in Figure 6(a) and (b), respectively. The average tax and cost is again computed within a time window of $T = 1000$ slots.

Similar to the two-SU network, SU 5 significantly reduces the packet loss rate by 14.6% and average cost by 16.1% by adopting the best response learning-based bidding strategy. Figure 6 (a) and (b) further verify the improvement of the performance for SU 1. However, other SUs' performances are decreased, as they need now to compete against a learning SU (i.e. SU 5), which is able to make better bids for the available resources.

Table 3. Performance of SU 1~5 with various bidding strategies in the five SUs network

| | SU 1 | | SU 2 | | SU 3 | | SU 4 | | SU 5 | |
|---|---|---|---|---|---|---|---|---|---|---|
| | Packet Loss Rate (%) | Average cost | Packet Loss Rate (%) | Averag e cost | Packet Loss Rate (%) | Average cost | Packet Loss Rate (%) | Averag e cost | Packet Loss Rate (%) | Averag e cost |
| 1 | 21.14 | 1.2002 | 19.99 | 1.1666 | 22.05 | 1.2123 | 21.37 | 1.1949 | 24.17 | 1.3101 |
| 2 | 25.03 | 1.2992 | 24.20 | 1.2993 | 25.72 | 1.3338 | 26.02 | 1.3568 | 9.56 | 1.0988 |

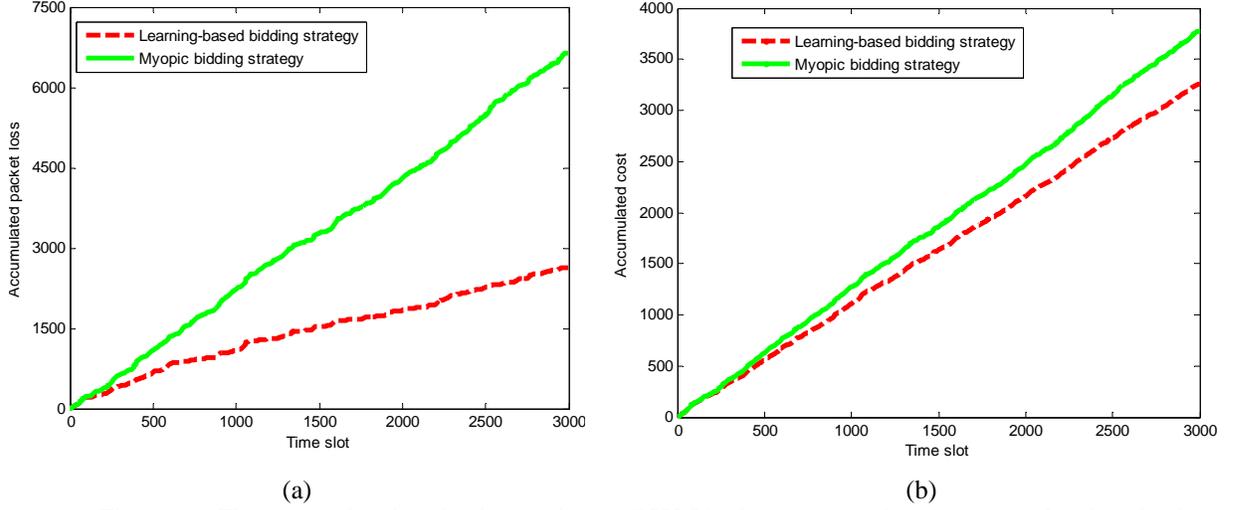

(a)                                                                  (b)

Figure 6.   The accumulated packet loss and cost of SU 5 in the two scenarios, (a) accumulated packet loss over the time slot; (b) accumulated cost over the time slot

## B. Impact of various dynamics on learning

In Section V.A, we demonstrate that the best response learning algorithm improves the bidding strategy, thereby leading to a reduced packet loss rate and average cost. In this simulation, we further investigate how various dynamics impact the learning algorithm proposed in Section IV.D. Specifically, we compare the learning performance under different channel dynamics, i.e. various available spectrum opportunities for the SUs as discussed in Section III.B. The source characteristics and channel conditions experienced by the SUs are kept the same as in Section V.A(1). We consider three types of channel dynamics corresponding to scenarios 1~3. The transition probabilities of the channel availability for all three scenarios are listed in Table 4. In each scenario, we compare two cases: in the first one, both SUs deploy myopic bidding strategies, and in the second one, SU 1 deploys best response learning-based bidding strategy, while SU 2 still uses the myopic bidding strategy.

Table 5 shows the average packet loss rate and cost experienced by the SUs under various channel dynamics. Interestingly, we observe from these results that even though the learning algorithm reduces the packet loss rate, it does not reduce the cost associated with SU 1, when the channel resources are abundant as in scenario 1. As the resources become increasingly scarce, the learning algorithm helps SU 1 to simultaneously reduce the packet loss rate and cost, e.g. in scenario 2 and 3. This observation can be explained as follows: when the resources are abundant, the cost (including the packet loss and tax) is small, i.e. the "value" of the channel is limited, and hence, the learning-based bidding strategy does not significantly benefit. On the other hand, when the resources are scarce, the bid vectors of the SUs in the current time slot will significantly affect the transition of their states through the channel allocation comparing to the case when the resources are abundant. For example, if an SU makes low bids as compared to other SUs, it might have no resources (channels) allocated to it when resources are scarce (i.e. the cognitive radio network is congested). In this case, the learning-based bidding strategy will carefully plan the bid by considering the future impact and thus, it is able to successfully improve the performance of SU 1 in terms of reducing the average cost.

Table 4. Channel availability probability

|  | Channel 1 | | | Channel 2 | | |
|---|---|---|---|---|---|---|
|  | $p_1^{NF}$ | $p_1^{FN}$ | Number of opportunities | $p_2^{NF}$ | $p_2^{FN}$ | Number of opportunities |
| Scenario 1 | 0.8 | 0.2 | 3502 | 0.8 | 0.2 | 3498 |
| Scenario 2 | 0.5 | 0.5 | 2490 | 0.5 | 0.5 | 2462 |
| Scenario 3 | 0.4 | 0.6 | 1960 | 0.4 | 0.6 | 1968 |

Table 5. Average packet loss rate and cost for the SUs under various resource constraints

|  |  | SU 1 | | SU 2 | |
|---|---|---|---|---|---|
|  |  | Packet loss rate | Average cost | Packet loss rate | Average cost |
| Scenario 1 | $\pi_1^{myopic}, \pi_2^{myopic}$ | 3.08 | 0.2678 | 2.90 | 0.2844 |
|  | $\pi_1^{\mathcal{L}}, \pi_2^{myopic}$ | 2.69 | 0.3092 | 4.17 | 0.4110 |
| Scenario 2 | $\pi_1^{myopic}, \pi_2^{myopic}$ | 21.36 | 1.8954 | 23.85 | 1.7471 |
|  | $\pi_1^{\mathcal{L}}, \pi_2^{myopic}$ | 14.54 | 1.6764 | 30.67 | 2.1744 |
| Scenario 3 | $\pi_1^{myopic}, \pi_2^{myopic}$ | 45.01 | 3.6283 | 45.42 | 3.8289 |
|  | $\pi_1^{\mathcal{L}}, \pi_2^{myopic}$ | 35.21 | 3.2590 | 56.44 | 4.5162 |

## VI. CONCLUSIONS

In this paper we model the cognitive radio resource allocation problem as a "stochastic game" played among strategic SUs. At each stage of the game, the CSM deploys a generalized second price auction mechanism to allocate the available spectrum resource. The SUs are allowed to simultaneously and independently make bid decision on that resource by considering their current states, experienced environment as well as the estimated future reward. To improve the bid decision at each stage, we propose a best response learning algorithm to predict the possible future reward at each state. The simulation results show that our proposed learning algorithm can significantly improve the SUs' performance. Our future work will focus on analyzing the performance of cognitive radio networks where multiple SUs are deploying various learning strategies and protocols.